\documentclass[sigconf]{acmart}
\usepackage{amssymb,amsmath}
\usepackage{multirow}
\usepackage{notoccite}

\AtBeginDocument{%
  \providecommand\BibTeX{{%
    \normalfont B\kern-0.5em{\scshape i\kern-0.25em b}\kern-0.8em\TeX}}}

\setcopyright{acmcopyright}
\copyrightyear{2020}
\acmYear{2020}
\acmDOI{XX.XXXX/XXXXXXX.XXXXXXX}

\acmConference[PETRA '20]{13th International Conference on Pervasive Technologies Related to Assistive Environments}{June 30 - July 3, 2020}{Corfu, Greece}
\acmPrice{15.00}
\acmISBN{978-1-4503-7773-7}

\begin{document}

\title{Mapping Motor Cortex Stimulation to Muscle Responses: A Deep Neural Network Modeling Approach}

\author{Md Navid Akbar}
\email{makbar@ece.neu.edu}
\orcid{0000-0003-3000-315X}
\affiliation{%
  \institution{Northeastern University}
  \city{Boston}
  \state{MA}
  \country{USA}
  \postcode{02115}
}

\author{Mathew Yarossi}
\email{m.yarossi@northeastern.edu}
\affiliation{%
  \institution{Northeastern University}
  \city{Boston}
  \state{MA}
  \country{USA}
}

\author{Marc Martinez-Gost}
\email{marc.martinez.gost@estudiant.upc.edu}
\affiliation{%
  \institution{Technical University of Catalonia}
  \city{Barcelona}
  \country{USA}
}

\author{Marc A. Sommer}
\email{marc.sommer@duke.edu}
\affiliation{%
 \institution{Duke University}
 \city{Durham}
 \state{NC}
 \country{USA}}

\author{Moritz Dannhauer}
\email{moritz.dannhauer@duke.edu}
\affiliation{%
 \institution{Duke University}
 \city{Durham}
 \state{NC}
 \country{USA}}

\author{Sumientra Rampersad}
\email{s.rampersad@northeastern.edu}
\affiliation{%
  \institution{Northeastern University}
  \city{Boston}
  \state{MA}
  \country{USA}
}

\author{Dana Brooks}
\email{brooks@ece.neu.edu}
\affiliation{%
  \institution{Northeastern University}
  \city{Boston}
  \state{MA}
  \country{USA}
}

\author{Eugene Tunik}
\email{e.tunik@neu.edu}
\affiliation{%
  \institution{Northeastern University}
  \city{Boston}
  \state{MA}
  \country{USA}
}

\author{\text{Deniz Erdo{\u{g}}mu{\c{s}}}}
\email{erdogmus@ece.neu.edu}
\affiliation{%
  \institution{Northeastern University}
  \city{Boston}
  \state{MA}
  \country{USA}
}

\renewcommand{\shortauthors}{Akbar, et al.}

%
\begin{abstract}
A deep neural network (DNN) that can reliably model muscle responses from corresponding brain stimulation has the potential to increase knowledge of coordinated motor control for numerous basic science and applied use cases. Such cases include the understanding of abnormal movement patterns due to neurological injury from stroke, and stimulation based interventions for neurological recovery such as paired associative stimulation. 
In this work, potential DNN models are explored and the one with the minimum squared errors is recommended for the optimal performance of the M2M-Net, a network that maps transcranial magnetic stimulation of the motor cortex to corresponding muscle responses, using: a finite element simulation, an empirical neural response profile, a convolutional autoencoder, a separate deep network mapper, and recordings of multi-muscle activation. We discuss the rationale behind the different modeling approaches and architectures, and contrast their results.
Additionally, to obtain a comparative insight of the trade-off between complexity and performance analysis, we explore different techniques, including the extension of two classical information criteria for M2M-Net.
Finally, we find that the model analogous to mapping the motor cortex stimulation to a combination of direct and synergistic connection to the muscles performs the best, when the neural response profile is used at the input.
\end{abstract}

\ccsdesc[500]{Computing methodologies~Neural networks}
\ccsdesc[100]{Computing methodologies~Supervised learning by regression}
\ccsdesc[100]{Computing methodologies~Cross-validation}
\ccsdesc[300]{Computing methodologies~Model verification and validation}
\ccsdesc[100]{Hardware~Neural systems}

\keywords{Transcranial magnetic stimulation (TMS), muscle synergy, convolutional neural network (CNN), performance-complexity analysis, autoencoder.}

\maketitle

\section{INTRODUCTION}
\label{sec:intro}
Coordinated activation of groups of muscles for voluntary movement is a prominent feature of human motor control \cite{bizzi2013neural}. Abnormal coordination of muscles is a marker of physical impairment in numerous congenital and acquired disease states such as cerebral palsy, stroke, and Parkinson’s disease. To optimize the use of neurotechnology and rehabilitation to improve impaired function, we must first gain a better understanding of cortical control of  healthy multi-muscle coordination \cite{ni2015transcranial}.
Transcranial magentic stimulation (TMS) describes a non-invasive procedure that uses magnetic fields to stimulate nerve cells in the brain \cite{hallett2007transcranial}. When applied to the motor cortex, TMS produces motor evoked potentials (MEPs) in a spatially selective set of muscles that can be recorded using standard surface electromyography (EMG). TMS, therefore, provides the best non-invasive approach to stimulation-recording for the study of motor control in humans. A deeper mechanistic understanding of the extent to which TMS can induce coordinated muscle activation is essential for innovation in TMS use as biomarker of disease or a means of intervention. For example, use of TMS as a diagnostic tool to identify cortical motor topography associated with abnormal muscle patterns may help identify who may benefit most from specific interventions following stroke, and track cortical changes associated with recovery \cite{yarossi2019association}. Used as a tool for neurological intervention, TMS synchronized with electrical peripheral nerve stimulation, has been demonstrated to strengthen neural connection and induce partial restoration of muscle activity \cite{shulga2016long}. Thus, TMS has the ability to produce positive plastic changes in the corticospinal tract of stroke-induced paralysis patients \cite{kubis2016non, hoyer2011understanding}. Such interventions could possibly be enhanced if precise TMS coil placement to stimulate a specific muscle group could be paired with multi-muscle electrical stimulation to enhance the functional grouping of muscles. A major challenge to both of the aforementioned use cases is to reliably locate the region of interest in the brain's motor cortex that maps to the desired muscle grouping. To address this challenge, we introduce deep learning in the modeling scenario.

Convolutional neural networks (CNNs) have already been demonstrated to outperform classical methods, such as support vector machines, for classification of EMG signals \cite{atzori2016deep}. Moreover, a CNN autoencoder has the capacity to successfully learn biologically plausible features \cite{masci2011stacked}. 
Our group has also previously described a preliminary framework for investigating cortical control of multi-muscle activation using TMS and deep learning, using a single stimulus intensity. 

In this work, we first present the mathematical model of our system. 
Subsequently, we methodically investigate different CNN architectures and a rigorous hyper-parameter fitting scheme to create a model that is more broadly generalizable, incorporating trials from four different stimulus intensities. 
We call this model M2M-Net (motor cortex to muscle network).
We then recommend the best model among those tested, for M2M-Net, and discuss the likely reasons governing the differences in performances. 
Finally, for assessing the trade-off between complexity and performance, we explore different statistical techniques, including extension of the Bayesian information criteria (BIC) and the Akaike information criteria corrected (AICc) for M2M-Net.

\section{EXPERIMENTAL METHODS}
\label{sec:exp}
\subsection{Data Acquisition} 
A 30 year old, right-handed, healthy male, eligible for TMS, participated following informed consent. Prior to TMS, a T1-weighted image (TI=1100ms, TE=2.63ms, TR=2000ms, 256x192x160 acquisition matrix, FOV=256x192mm, 1mm$^3$ voxels) was used for neuronavigation (Brainsight, Rogue Research). The procedure used for TMS mapping has been described by our group elsewhere \cite{yarossi2019experimental}. Briefly, the subject was seated, with forearms supported. Surface EMG (Trigno, Delsys, 2kHz) was recorded from 15 hand-arm muscles during TMS.
The TMS coil (Magstim 200, 70mm figure-8) was held tangential to the scalp with the handle posterior and 45$^\circ$ to midline. The right 1st dorsal interosseus (FDI) muscle hotspot was found via a coarse map of the hand knob area. Peak-to-peak EMG amplitude 20-50msec after the TMS pulse was the outcome variable (Matlab, The Mathworks). TMS intensity during mapping in proportion to the resting motor threshold (RMT), which was the minimum intensity required to elicit MEPs $>$50 $\mu$V on 3/6 consecutive trials. TMS (100-300 stimuli, 4sec ISI) was delivered over a 7cm$^{2}$ area centered on the hotspot using stimulus intensities of 110\%, 120\%, 130\% and 140\% of RMT. For each intensity, one stimulus was delivered to each of 49 equidistant grid points, and the remaining stimuli were delivered using real time feedback of MEPs to maximize information about responsive areas \cite{niskanen2010group}. 
\subsection{Model Components}
We have previously described a forward model comprised of four components to predict TMS-evoked multi-muscle activation based on stimulus parameters alone \cite{yarossi2019experimental}.  

\begin{enumerate}
  \item A finite element (FE) model that generates, from TMS parameters (coil geometry/position/orientation, magnetic pulse characterization, stimulus intensity), a volumetric E-field on a mesh that accurately represents tissue segmentation of the subject specific MRI \cite{Brainstimulator,SCIRun}.
   
    \item A nonlinear mapping of local E-fields to the expected neural ensemble firing rates, based on nonhuman primate experimental data. This transformation uses a sigmoidal fit, to model the dose-response curve for putative excitatory neurons that may project to spinal circuits \cite{mueller2014simultaneous,grigsby2015neural}.
    
    \item A two-stage neural network model we call M2M-Net that maps either Step 1 or Step 2 component outputs to multi-muscle responses. Model selection for the M2M-Net is the main subject of this paper.
    
    \item A low dimensional linear structure,  which we refer to as Synergy (to contrast with MEP), representing synergistic muscle activation in coherent groups,  extracted using non-negative matrix factorization \cite{tresch2006matrix,lee1999learning} applied to recorded MEPs. Essentially, they function as a low-rank non-negative approximation of the MEP data. 
\end{enumerate}  

\begin{figure}[hbt]
\includegraphics[width=\columnwidth]{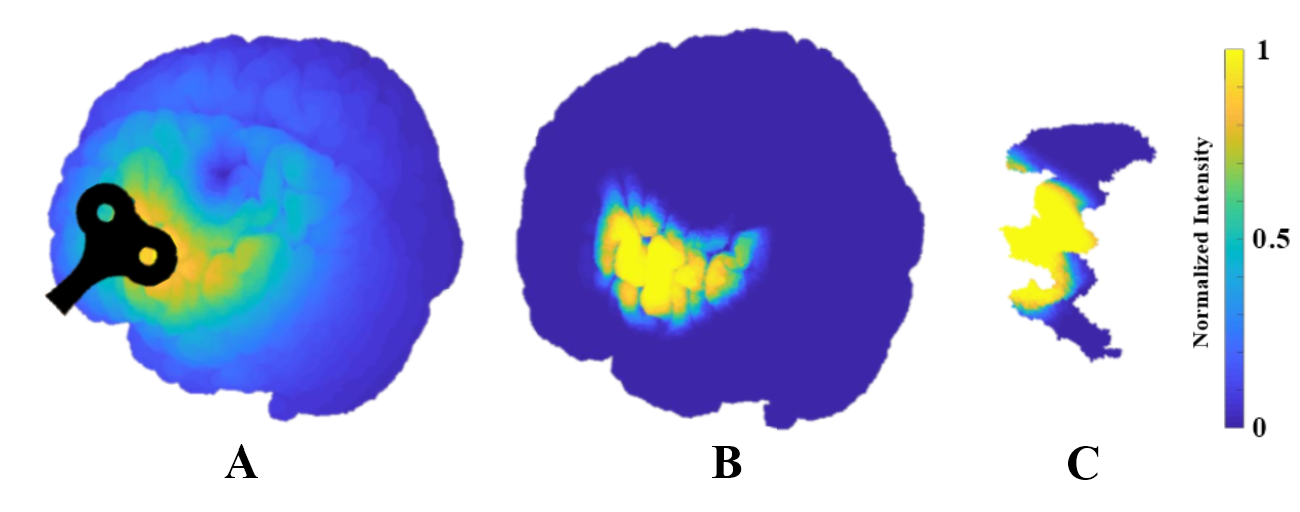}
\centering
\caption{Normalized E-field distributions in the brain. (A) The E-field generated on the brain as a result of stimulation by the TMS coil. (B) The high contrast intensity observed after application of the E-field to neural firing transfer function. (C) Mask of the motor cortex.}
\label{fig_e_field_brain}
\end{figure}

Fig.~\ref{fig_e_field_brain} illustrates Steps 1 and 2.

\subsection{System Model}

\label{sec:model}
The mathematical model of the system for mapping directly to the MEPs (Direct) is given by
\begin{equation}
    \Vec{m}_i = f_D(\mathbf{X}_i) + \Vec{\epsilon}_{D,i}, \label{eq_direct_model}
\end{equation}
where $\Vec{m_i}$ is the $k \times 1$ ($k$ = 15 muscles)  observed muscle activity vector for the $i$th trial, $\mathbf{X}_i \in \mathbb{R}^{l \times l \times l}$ ($l$ = 64) is the input 3D matrix corresponding to the E-field distribution or the neural firing rate, $f_D()$ represents M2M-Net for the Direct connection, $\Vec{\epsilon}_{D,i}$ is the $k \times 1$ residual mapping error, and $i \in 1,2,...,N$ ($N$ = 629 is total number of trials for training, from the four different stimulus intensities).

\begin{figure*}[htb]
\includegraphics[width=\textwidth]{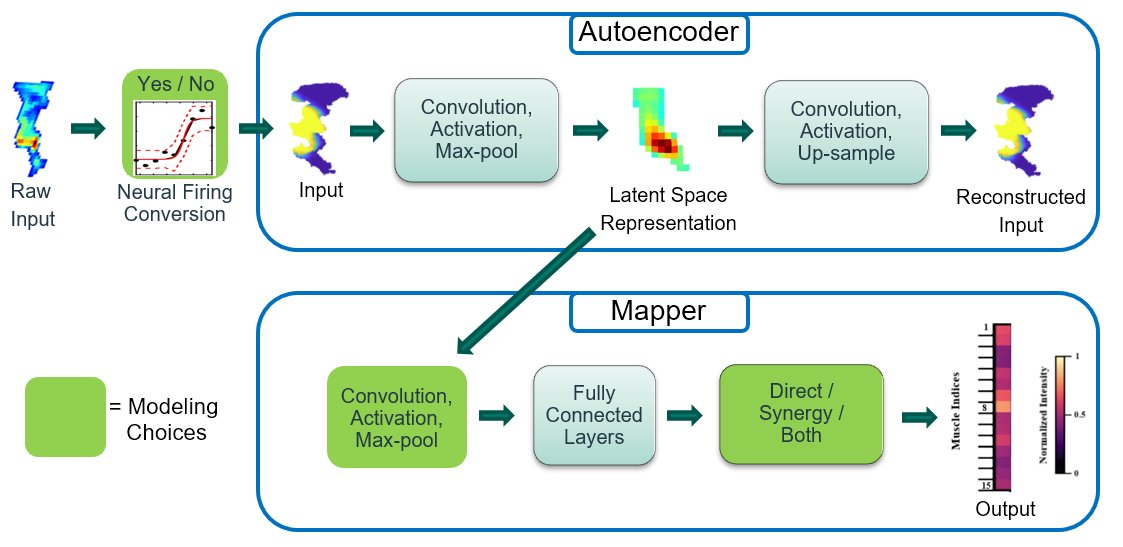}
\caption{Block diagram of the system model.}
\label{fig_sys_mod}
\end{figure*}

In the case of mapping to the synergies (Synergy), the muscle activation vector can also be expressed as follows
\begin{equation}
    \Vec{m}_{i} = \Tilde{\mathbf{B}}\Vec{a_i} + \Vec{w_i}, \label{eq_synergy_base}
\end{equation}
where $\Tilde{\mathbf{B}} \in \mathbb{R}^{k \times r}$ ($r$ = 9 synergies) is the unknown basis matrix, in which each row indicates the relative activation level of each muscle in all the synergies, $\mathbf{a}_i$ is the $r \times 1$ unknown activation vector that linearly combines synergy bases, and $\mathbf{w}_i \sim\mathcal{N}(\mathbf{0},\sigma^2\mathbf{I})$ is a spatiotemporal white additive Gaussian noise.
Using the maximum likelihood (ML) criteria and non-negative matrix factorization (NMF) technique as described in \cite{yarossi2019experimental}, we estimate $\mathbf{B}$ to give the following relation
\begin{equation}
    \mathbf{M}  \approx \mathbf{B} \mathbf{A}, \label{eq_NMF}
\end{equation}
where $\mathbf{M} \in \mathbb{R}^{k \times N}$ is the matrix of the all the muscle activations given by $\mathbf{M} = \big[ \Vec{m}_{1} \: \: \Vec{m}_{2} \dots \Vec{m}_{N} \big]$, and $\mathbf{A} \in \mathbb{R}^{r \times N}$ is the matrix of all the activation vectors given by $\mathbf{A} = \big[ \Vec{a}_{1} \: \: \Vec{a}_{2} \dots \Vec{a}_{N} \big]$.
Finally, our Synergy model takes the form
\begin{equation}
    \Vec{m}_i = \mathbf{B} f_S (\mathbf{X}_i) + \Vec{\epsilon}_{S,i}. \label{eq_synergy_model}
\end{equation}

For the third case, we first train M2M-Net on the activation matrix representing the synergies, given by $\mathbf{A}$. Afterwards, we train it on the matrix $\mathbf{\xi}_S \in \mathbb{R}^{k \times N}$ of all the residual errors from the Synergy model given by $\mathbf{\xi}_S = \big[ \Vec{\epsilon}_{S,1} \: \: \Vec{\epsilon}_{S,2} \dots \Vec{\epsilon}_{S,N} \big]$, using the Direct connection. This is given by
\begin{equation}
    \Vec{\epsilon}_{S,i} = f_{D^*} (\mathbf{X}_i) + \Vec{\epsilon}_{B,i}. \label{eq_error_map}
\end{equation}
Lastly, the predictions obtained from the Synergy and the Direct connections are then added to give the combinatorial model (Both)
\begin{equation}
    \Vec{m}_i = \mathbf{B} f_S (\mathbf{X}_i) + f_{D^*} (\mathbf{X}_i) + \Vec{\epsilon}_{B,i}. \label{eq_both_model}
\end{equation}

\subsection{Network Training and Testing Procedures}

M2M-Net is a hybrid deep network, composed of a convolutional autoencoder used for dimensionality reduction and a separate deep CNN mapper, as seen in Fig. \ref{fig_sys_mod}. For each stimulation, the 64-cube three-dimensional E-fields, or ensemble firing rates, serve as the input to the autoencoder.
The autoencoder provides low-dimensional, 16-cube representations of the input, which are then passed on to the mapper. The mapper contains a sequence of convolutional layers. The output from these layers are then flattened and connected to fully connected dense layers. The structure of the last dense layer determines if outputs will be mapped via Direct, via Synergy, or via Both. 

We tested two common techniques of preprocessing methods applied to both the E-field and neural firing rate inputs: mean standardization and min-max scaling. Both of these methods have been shown to speed up the training process, as well as improve prediction accuracy \cite{7808140}. For the output, activations for the individual muscles were min-max scaled to the unit interval [0,1].

Each set of inputs (E-field distributions or neural firing rates) and associated outputs (muscle activity or synergy activations) are split into calibration and test sets, in a 9:1 ratio. The calibration set is further split into training and validation sets. A 10-fold cross validation scheme is used to tune hyperparameters, within the calibration set. The lowest normalized root mean squared error (NRMSE) performance on the validation set determines the best choice for the hyperparamaters.
Once the choice of hyperparameters are decided, a total of 12 different models are trained using the entire calibration set, and their respective
performances are assessed on the testing set. This process is repeated three times, each time randomly selecting different subsets of the entire data for calibration and test. The NRMSE of a test set of length $T$ ($T$ = 70) is calculated as
\begin{equation}
    \text{NRMSE} = \sqrt{\frac{\sum_{j=1}^{T} \| \Vec{\epsilon}_j \|^2_2} {\sum_{j=1}^{T} \|\Vec{m}_{j}\|^2_2}}, \label{eq_NRMSE}
\end{equation}
where $\|.\|_2$ denotes the Euclidean norm.

During training, the inputs are processed in mini-batches of size 32.
Stochastic gradient descent with momentum is selected as the optimizer for the autoencoder, while Adadelta is chosen for the mapper. A dynamic reduction in learning rate is scheduled with the increase in epoch, for each optimizer.  
Further details of the network layers and parameters are outlined in Section \ref{choice_of_parameters}.

We select NRMSE as the performance evaluation metric for our 12 models.
Later, we will also illustrate the variation of NRMSE with the number of parameters in the mapper, for the models of interest. 
However, to attain a more meaningful insight, we will also attempt to extend two classical statistical techniques: the Bayesian information criteria (BIC) and the Akaike information criteria corrected (AICc), for M2M-Net.

Following the model outlined in \cite{stoica2004model}, our muscle activation vector may be represented in the form
\begin{equation}
    \Vec{m}_i = \mu (\Vec{\gamma}_i) + \Vec{\epsilon}_{i}, \label{eq_GSN_model}
\end{equation}
where $\Vec{\gamma}_i$ is an unknown parameter vector, and $\Vec{\epsilon}_{i}$ is now assumed to be Gaussian with mean zero and covariance matrix given by $E\{ee^{\intercal}\} = \sigma^2 \mathbf{I}$.
The ML estimate of $\sigma^2$ for $N$ trials then becomes
\begin{equation}
    \hat{\sigma}^2 = \frac{1}{Nk} \sum_{i=1}^{N} \| \Vec{m}_i - \mu (\Vec{\gamma}_i) \|^2_2 . \label{eq_GSN_model}
\end{equation}
The corresponding value of log-likelihood of the probability density function of $\mathbf{M}$ then becomes $(Nk \log \hat{\sigma}^2 + constant)$.
For BIC and AICc, each expression includes this value of the log-likelihood (the constant is trivial) plus a characteristic penalty term involving the number of free parameters, $p$, to be estimated (in the mapper) \cite{stoica2004model, hurvich1989regression}.
We can thus express
\begin{equation}
    \text{BIC} = (Nk) \log \hat{\sigma}^2 + p \log(Nk), \label{eq_BIC}
\end{equation}
and
\begin{equation}
    \text{AICc} = (Nk) \log \hat{\sigma}^2 + \frac{Nk(p+Nk)}{Nk-p-2}. \label{eq_AICc}
\end{equation}

\subsection{Layers and Parameters}
\label{choice_of_parameters}
In this section, we describe the methodology of parameter selection for our network shown in Fig. \ref{fig_sys_mod}.

\underline{Convolution Layers}: M2M-Net contains two sets of convolution-activation-maxpool layers in the encoder portion of the autoencoder, and three such sets in the 3-layer mapper. For the 6-layer mapper, two identical convolution layers are used before each activation layer. Going back to the autoencoder, there are two sets of symmetric convolution-activation-upsample layers in its decoder portion.
The number of filters in subsequent convolution layers are doubled: from 32 to 64 for the encoder of the autoencoder, and from 16 to 64 for the mapper. For the decoder, it is reversed.
This choice is aligned with the established understanding that while low-level filters learn basic shapes, high-level filters are capable of identifying more sophisticated patterns \cite{simonyan2014very}. The convolutional layers in M2M-Net have 3$\times$3$\times$3 filters, except the ones used to prepare the intermediate 16-cube representation and the final 64-cube reconstruction. Both of these layers have a 1$\times$1$\times$1 filter window, and a single channel.
The padding used is 1 element on each side, and the stride is 1.

\underline{Activation layers:} Activation functions follow all convolutional and dense layers. The exponential linear unit (eLU) is the activation function of choice, 
after every convolutional layer in the autoencoder. However, for the mapper, the rectified linear unit (ReLU) is chosen as it outperformed the eLU in our cross validation rounds. The final activation function in the autoencoder is a linear function. In the mapper, it is a sigmoid for connections to the MEPs (Direct), or that to the synergies (Synergy): since the output should be constrained between 0 and 1. When mapping to the differences between the synergy activations and the ground truth via Both, the result could contain negative values as well. Thus, the second Direct connection to the residuals, indicated earlier in \eqref{eq_error_map}, have linear activations in the intermediate layers, and a tanh as the final activation, to constraint the output between -1 and 1.

\begin{table*}[ht]
\begin{center}
\caption{NRMSE values across different models.}
\begin{tabular}{|c|c|c|c|c|c|c|}
    \hline
    & \multicolumn{3}{c|}{3 conv layer Mapper } & \multicolumn{3}{c|}{6 conv layer Mapper}\\
    \hline
    & Direct & Synergy & Both & Direct & Synergy & Both\\
    \hline
    Raw E-field Input & 0.658$\pm$0.042  & 0.671$\pm$0.049  & 0.661$\pm$0.048 & 0.644$\pm$0.058 & 0.638$\pm$0.053 & 0.630$\pm$0.056 \\
    \hline 
    Neural Firing Filter & 0.519$\pm$0.048 & 0.505$\pm$0.041 & \textbf{0.500$\pm$0.043} & 0.518$\pm$0.045 & 0.516$\pm$0.049 & 0.508$\pm$0.049 \\
    \hline 
\end{tabular}
\label{tab_1}
\end{center}
\end{table*}

\underline{Pooling, upsampling, and dense layers:} Max-pooling and upsampling layers are used to reduce and increase the sizes of the representations, respectively. All filter windows for these layers have a common size of 2$\times$2$\times$2, and a stride of 1. 
The final multi-dimensional tensor in the mapper is flattened to a one-dimensional vector, which is then fully connected to the dense layers.

\underline{Regularization:} Each convolutional layer in both network is fitted with $\ell_1$ weight regularization. Our input E-field distribution data are sparse.
It is a well known technique to use $\ell_1$ penalty in such convolutional sparse coding \cite{bristow2013fast}. Batch normalization layers are used to prevent internal covariance shifts in the data \cite{ioffe2015batch}, and follow the convolution layers in the mapper.
Each dense layer is followed by a dropout layer, which helps counter against over-fitting \cite{srivastava2014dropout}. The values of the hyper-parameters for both the dropout layers and the $\ell_1$ penalty are chosen from the cross validation rounds performed within the calibration set, and are found to be 0.15 and 1e-4, respectively.


\section{RESULTS}
\label{sec:results}

From our experiments, min-max scaling across the entire input dataset proved to be the better of the two preprocessing techniques. It provided more distinct intermediate representations and lowered prediction errors. This scaling was performed once again on the entire set of intermediate representations, before they form input to the mapper.



Table \ref{tab_1} presents the arithmetic means and standard deviations of the performance evaluations for the different models. 
Consistent with our previous results from \cite{yarossi2019experimental} for the TMS intensity at 110\% RMT, we saw that the generalized model across four different intensities also performed better with the E-field to neural firing rate conversion, compared to E-fields used directly as input. This could be explained in light of the neural firing rate conversion effectively thresholding and increasing the image contrast. Since the images are now more focused in the regions of interest, feature extraction becomes easier, and the model performs better. Examining different mapper combinations corresponding to neural firing, we observe that the prediction by mapping to Synergy outperformed mapping via Direct. Model accuracy was the greatest using the Both model.
We also observed that the 3 layer mapper performed better compared to the 6 layer mapper. This could be a case of model complexity and overfitting, as the 6 layer models have more trainable parameters, compared to their 3 layer counterparts. Results from one experiment, out of the three conducted, is shown for the 3 layer mapper in Fig. 3.  
\begin{figure}[hbt]
\begin{center}
\includegraphics[width=\columnwidth]{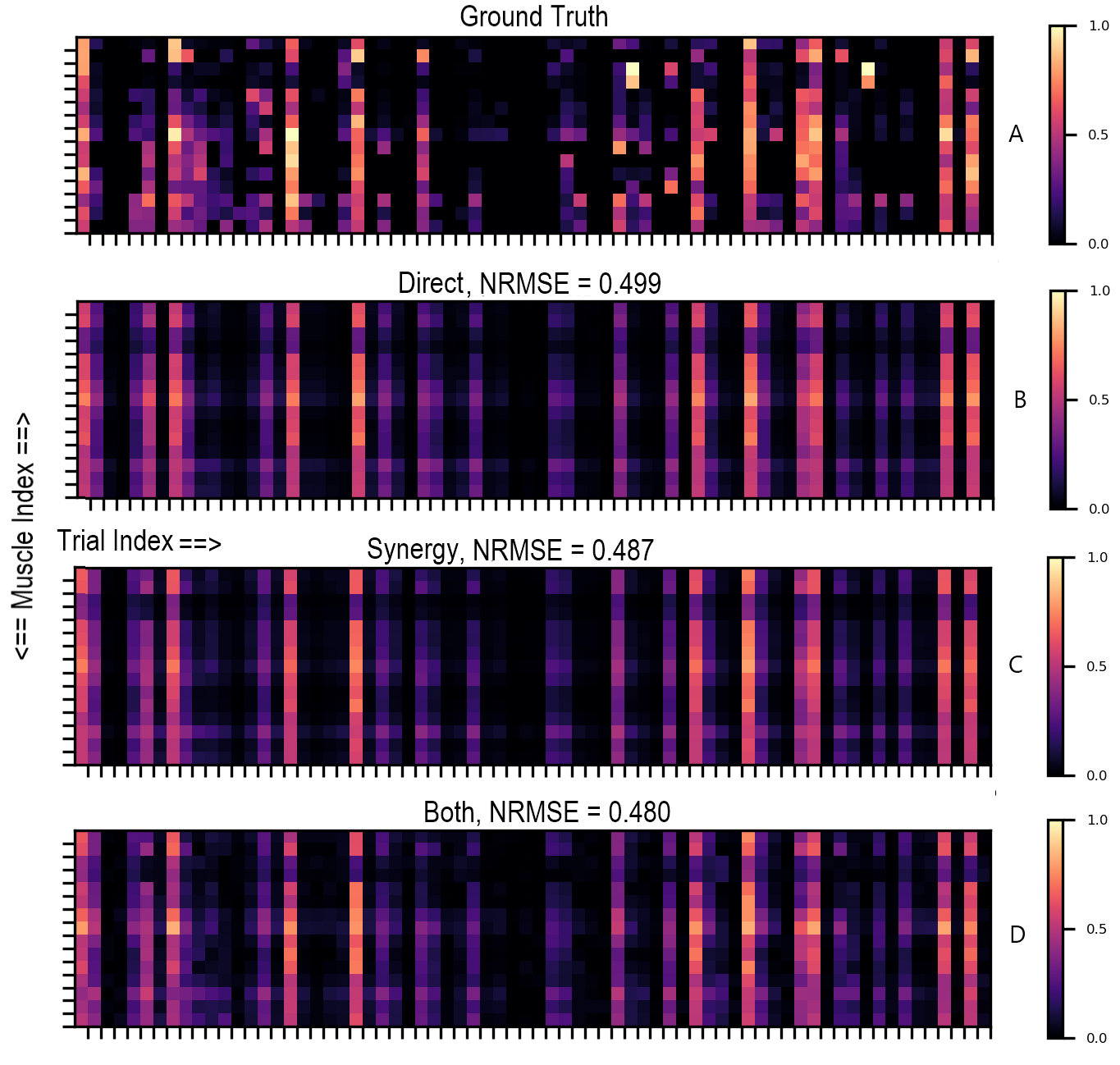}
\caption{Predicted muscle activation for the 3-layer mapper with the neuron firing input on a test set, for 15 muscles and 70 trials. (A) Ground truth. (B) Prediction with Direct. (C) Prediction with Synergies. (D) Prediction with Both.}
\end{center}
\label{fig_prediction_sample}
\end{figure}

With the models using the E-fields directly as input, we notice that the 6 layer models outperform their corresponding 3 layer counterparts. This time, since no contrast enhancing transfer function was used for these models, the network needed to identify the important zones entirely by itself. A model with more degrees of freedom, related to its number of free parameters \cite{gao2016degrees}, would be inclined to perform better in such a scenario, similar to what is observed. Hence, the more complex models provide the better predictions.

When comparing Fig. 3 and the row corresponding to the neural firing input in Table \ref{tab_1}, we note that the means and the deviations appear rather high. This could be explained if we take into account the many stimulations that produce null or low muscle activations. In accordance to \eqref{eq_NRMSE}, we realize that even if the predictions by M2M-Net result in a small $\Vec{\epsilon}_j$, the corresponding $\Vec{m}_j$ is smaller, or even zero. As a consequence, the overall NRMSE goes up. The same factor plays a role in the deviation values. Depending on whether or not a large number of such stimulations end up in our random selections for the test sets, the NRMSE value varies, and hence a high deviation. Interesting to note though, in each of our test runs, the relative performances among the different models remained almost exactly the same. This indicates that even though the deviations might hint a cause of concern, the relative behavior of M2M-Net for the different models is preserved across the different runs.

\begin{figure}[htb]
\begin{center}
\includegraphics[width=\columnwidth]{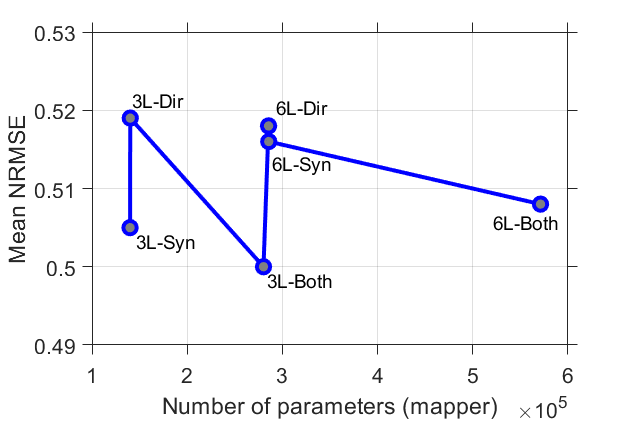}
\caption{Performance vs. complexity for the models with the neuron firing input.}
\end{center}
\label{fig_bic}
\end{figure}

In Fig. 4, we can see the interplay between performance and complexity of the models with the neural firing input.
As observed, the error reaches a minimum with the 3-layer mapper connected via Both, with about $2.8 \times 10^5$ parameters. Below this, the number of parameters are probably insufficient to model the complexity of the network. Above it, there are probably too many parameters, and those models likely overfit. Of the models we tested, a 3-layer mapper connected via Both probably has the optimum number of parameters.

To explore the trade-off between performance and complexity further, Table 2 lists the BIC and AICc values for the models indicated. We notice that both criteria select the 3-layer mapper with the neuron firing input as the model with the lowest score, indicating it to have the best performance-complexity value. Interestingly, they differ in the choice of the second best model, and so on. Understandably, we have to interpret the values in Table 2 with caution.
CNNs are generally over-parameterized. Hence, 
classical statistical tools such as BIC and AICc would tend to favor the least complex model, at the cost of training performance. Again, training performance of a CNN is not necessarily a correct indicator of its test performance, or generalizability. 
On the other hand, it might bear useful insights to know how a model performs, taking into account (or penalizing) its complexity. 

\begin{table}[htb]
\begin{center}
\caption{BIC and AICc values for the models with the neuron firing input.}
\begin{tabular}{|c|c|c|c|c|c|c|}
    \hline
    & \multicolumn{3}{|c|}{3-layer Mapper } & \multicolumn{3}{c|}{6-layer Mapper}\\
    \hline
    & Dir & Syn & Both & Dir & Syn & Both\\
    \hline
    BIC ($10^6$) & 1.221  & 1.219  & 2.501 & 2.552 & 2.550 & 5.164 \\
    \hline
    AICc ($10^4$) & -7.266  & -7.320  & -7.269 & -7.257 & -7.239 & -7.220 \\
    \hline 
\end{tabular}
\label{tab_2}
\end{center}
\end{table}

\section{Conclusion}

In this work, we have systemically explored different CNN architectures, to develop a suitable model for mapping muscle responses corresponding to varying TMS intensities.
The model with the lowest NRMSE corresponded to the use of the E-field to neural firing rate conversion, a 3-layer mapper, and the use of both direct and synergy connections. This result demonstrates the value of the use of empirical knowledge at both ends of the network: in the form of firing rate conversion at the input, and the use of direct and synergistic connections at the output, reflecting corticospinal anatomy \cite{yarossi2019experimental}. Greater generalizability, with the inclusion of the synergy layer and using a lesser complex model, is indicative of the current theoretical understanding of the purpose of muscle synergies in the motor system \cite{bizzi2013neural}. Given the physiological plausibility of the model proposed, future work will concentrate on the validation of this model further and its eventual incorporation as a diagnostic utility or means of intervention for neurological impairment. 





\begin{acks}
This project was funded in part by grants R01-NS088674 (MS), P41
GM103545-18 (DB,MD), R01-NS085122 and R01-R01HD058301 (ET),
CMMI-193537 (ET, DE), CBET-1804550 (ET, DE, DB)
\end{acks}



\bibliographystyle{ACM-Reference-Format}
\bibliography{PETRA}

\end{document}